\documentclass{article}

\usepackage[square,sort,comma,numbers]{natbib}
\usepackage[preprint]{neurips_2019}

\usepackage[utf8]{inputenc} 
\usepackage[T1]{fontenc}    
\usepackage{hyperref}       
\usepackage{url}            
\usepackage{booktabs}       
\usepackage{amsfonts}       
\usepackage{nicefrac}       
\usepackage{microtype}      

\usepackage{graphicx}
\usepackage{amsmath}
\usepackage{amssymb}
\usepackage{float}
\usepackage{subfig}
\usepackage{wrapfig}
\usepackage{color}

\title{Fashion Editing with Adversarial Parsing Learning}

\author{%
  Haoye Dong$^1$, Xiaodan Liang$^1$, Yixuan Zhang$^2$, Xujie Zhang$^1$, \\
  \textbf{Zhenyu Xie$^1$, Bowen Wu$^1$, Ziqi Zhang$^1$, Xiaohui Shen$^3$, Jian Yin$^1$} \\
  $^1$Sun Yat-sen University, $^2$Petuum Inc, $^3$ByteDance AI Lab\\
  \texttt{donghy7@mail2.sysu.edu.cn, xdliang328@gmail.com} \\
}

\begin{document}

\maketitle

\begin{abstract}
   Interactive fashion image manipulation, which enables users to edit images with sketches and color strokes, is an interesting research problem with great application value. Existing works often treat it as a general inpainting task and do not fully leverage the semantic structural information in fashion images. Moreover, they directly utilize conventional convolution and normalization layers to restore the incomplete image, which tends to wash away the sketch and color information. In this paper, we propose a novel Fashion Editing Generative Adversarial Network (FE-GAN), which is capable of manipulating fashion images by free-form sketches and sparse color strokes. FE-GAN consists of two modules: 1) a free-form parsing network that learns to control the human parsing generation by manipulating sketch and color; 2) a parsing-aware inpainting network that renders detailed textures with semantic guidance from the human parsing map. A new attention normalization layer is further applied at multiple scales in the decoder of the inpainting network to enhance the quality of the synthesized image. Extensive experiments on high-resolution fashion image datasets demonstrate that the proposed method significantly outperforms the state-of-the-art methods on image manipulation. 
\end{abstract}

\section{Introduction}
\begin{figure}[ht]
    \centering
    \includegraphics[width=1.0\hsize \hspace{0.01\hsize}]{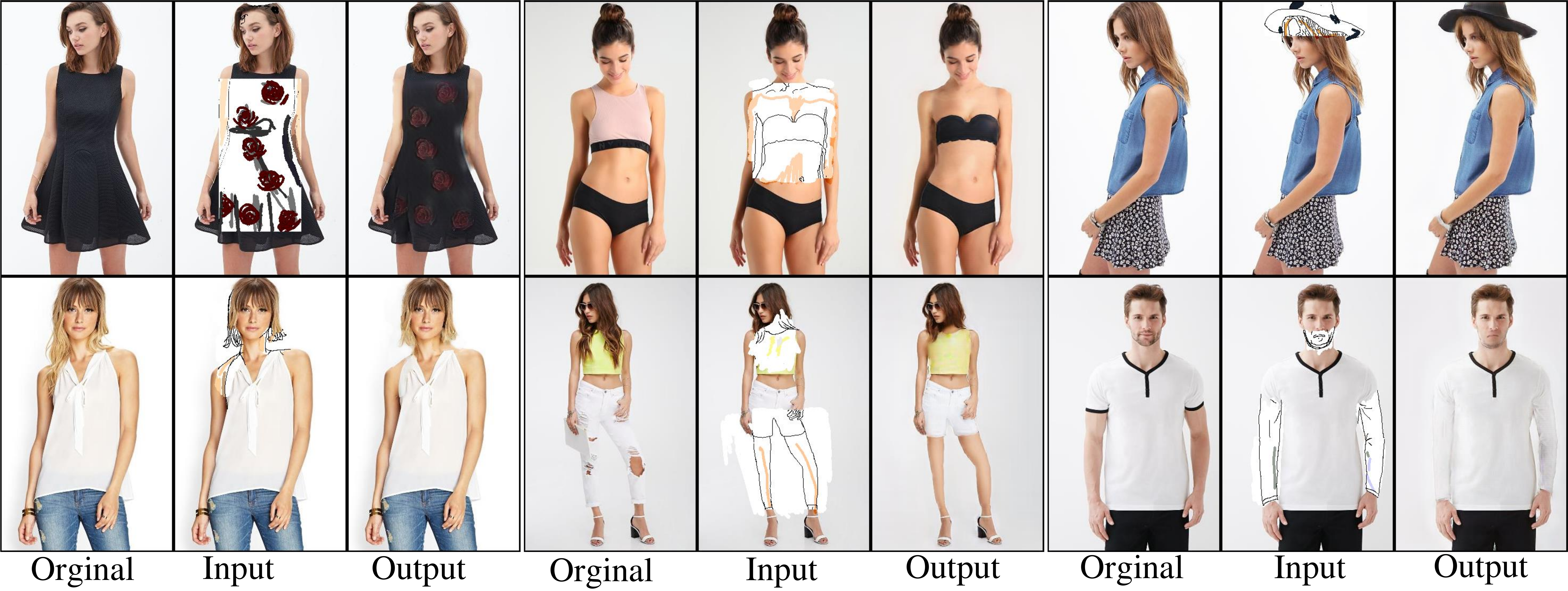} 
    \caption{Some interactive results of our FE-GAN. The input contains free-form mask, sketch, and sparse color strokes. Zoom in for details.}
    \label{fig:fig1}
\end{figure}

Fashion image manipulation aims to generate high-resolution realistic fashion images with user-provided sketches and color strokes. It has huge potential values in various applications. For example, a fashion designer can easily edit clothing designs with different styles; filmmakers can design characters by controlling the facial expression, hairstyle, and body shape of the actor or actress. In this paper, we propose FE-GAN, a fashion image manipulation network that enables flexible and efficient user interactions such as simple sketches and a few sparse color strokes. Some interactive manipulation results of FE-GAN are shown in Figure~\ref{fig:fig1}, which indicates that it can generate realistic images with convincing and desired details by controlling the sketch and color strokes.

In general, image manipulation has made great progress due to the significant improvement of neural network techniques~\cite{choi2017stargan, goodfellow2014generative,han2017viton,johnson2016perceptual,ma2017pose,ronn2015unet,zhu2017cycleGAN}. However, previous methods often treat it as an end-to-end one-stage image completion problem without flexible user interactions~\cite{yu2018Deepfillv2,Liu2018ImageIF,Nazeri2019EdgeConnectGI,pathak2016inpainting,Portenier2018faceshop,jo2019sc-fegan,Yu2018GenerativeII}. Those methods usually do not explicitly estimate and then leverage the semantic structural information in the image. Furthermore, they excessively use the conventional convolutional layers and batch normalization, which significantly dissolve the sketch and color information from the input during propagation. As a result, the generated images usually contain unrealistic artifacts and undesired textures.

To address the above challenges, we propose a novel Fashion Editing Generative Adversarial Network (FE-GAN), which consists of a free-form parsing network and a parsing-aware inpainting network with multi-scale attention normalization layers. Different from the previous methods, we do not directly generate the complete image in one stage. Instead, we first generate a complete parsing map from incomplete inputs, and then render detailed textures on the layout induced from the generated parsing map. Specifically, in the training stage, given an incomplete parsing map obtained from the image, a sketch, sparse color strokes, a binary mask, and a noise sampled from the Gaussian distribution, the free-form parsing network learns to reconstruct a complete human parsing map guided by the sketch and color. A parsing-aware inpainting network then takes the generated parsing map, the incomplete image, and composed masks as the input of encoders, and synthesizes the final edited image. To better capture the sketch and color information, we design an attention normalization layer, which is able to learn an attention map to select more effective features conditioned on the sketch and color. The attention normalization layer is inserted at multiple scales in the decoder of the inpainting network. Moreover, we develop a foreground-based partial convolutional encoder for the inpainting network that is only conditioned on the valid pixels of the foreground, to enable more accurate and efficient feature encoding from the image. 

We conduct experiments on our newly collected fashion dataset, named FashionE, and two challenging datasets: DeepFashion~\cite{liu2016deepfashion} and MPV~\cite{dong2019towards}. The results demonstrate that incorporating the multi-scale attention normalization layers and the free-form parsing network can help our FE-GAN significantly outperforms the state-of-the-art methods on image manipulation, both qualitatively and quantitatively. The main contributions are summarized as follows: 1) We propose a free-form parsing network that enables users to control parsing generation flexibly by manipulating the sketch and color. 2) We develop a newly attention normalization for extracting features effectively based on a learned attention map. 3) We design a parsing-aware inpainting network with foreground-aware partial convolutional layers and multi-scale attention normalization layers, which can generate high-resolution realistic edited fashion images.


\section{Related Work}
\textbf{Image Manipulation}. Image manipulation with Generative Adversarial Networks (GANs) \cite{goodfellow2014generative} is a popular topic in computer vision, which includes image translation, image completion, image editing, etc. Based on conditional GANs~\cite{mirza2014cgan}, Pix2Pix~\cite{isola2017pix2pix} is proposed for image-to-image translation. Targeting at synthesizing high-resolution photo-realistic image, Pix2PixHD~\cite{wang2017pix2pixHD} comes up with a novel framework with coarse-to-fine generators and multi-scale discriminators. ~\cite{Iizuka2017GloLoc, Yu2018GenerativeII} design frameworks to restore low-resolution images with an original (square) mask, which generate some artifacts when facing the free-form mask and do not allow image editing. To make up for these deficiencies, Deepfillv2~\cite{yu2018Deepfillv2} utilizes a user's sketch as input and introduces a free-form mask to replace the original mask. On top of Deepfillv2, Xiong et al.~\cite{Xiong2019ForegroundawareII} further investigate a foreground-aware image inpainting approach that disentangles structure inference and content completion explicitly.
 Faceshop~\cite{Portenier2018faceshop} is a face editing system that takes sketch and color as input. However, the synthesized image would have blurry edges on the restored region, and it would obtain undesirable result if too much area erased. Recently, another face editing system SC-FEGAN~\cite{jo2019sc-fegan} is proposed, which generates high-quality images when users provide the free-form as input. However, SC-FEGAN is designed for face editing. In this paper, we propose a novel fashion editing system conditioned on the sketch and sparse color, utilizing feature involved in the parsing map, which is usually ignored by previous methods. Besides, we introduce a novel multi-scale attention normalization to extract more significant features conditioned on the sketch and color.

\textbf{Normalization Layers}. Normalization layers have become an indispensable component in modern deep neural networks. Batch Normalization (BN) used in Inception-v2 network~\cite{ioffe2015batch}, making the training of deep neural networks easier. Other popular normalization layers, including Instance Normalization (IN)~\cite{uiyanov2016in}, Layer Normalization (LN)~\cite{ba2016ln}, Weight Normalization (WN)~\cite{salimans2016wn}, Group Normalization (GN)~\cite{wu2018gn}, are classified as unconditional normalization layers because no external data is utilized during normalization. In contrast to the above normalization techniques, conditional normalization layers require external data. Specifically, layer activations are first normalized to zero mean and unit deviation. Then a learned affine transformation is inferred from external data, which is utilized to modulate the activation to denormalized the normalized activations. The affine transformations are various among different tasks. For style transfer tasks~\cite{vincent2017cbn, hung2017cin}, affine parameters are spatially-invariant since they only control the global style of the output images. As for semantic image synthesis tasks, SPADE~\cite{park2019spade} applies a spatially-varying affine transformation to preserve the semantic information. In this paper, we propose a novel normalization technique named attention normalization. Instead of learning the affine transformation directly, attention normalization learns an attention map to extract significant information from the normalization activations. What's more, compared to the SPADE ResBlk in SPADE~\cite{park2019spade}, attention normalization has a more compact structure and occupies less computation resource.


\section{Fashion Editing}
\begin{figure}[t]
    \centering
    \includegraphics[width=1.0\hsize \hspace{0.01\hsize}]{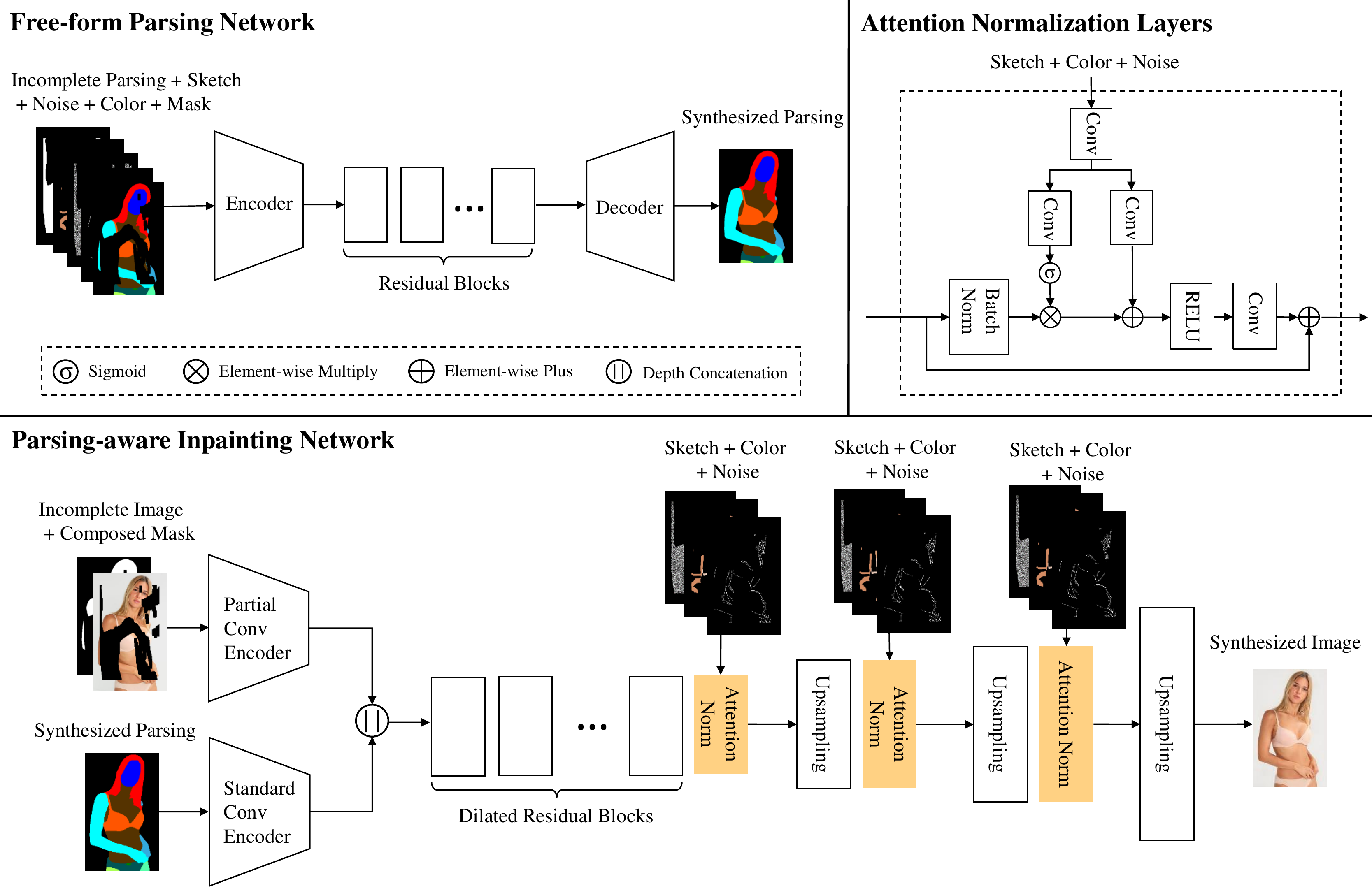} 
    \caption{The overview of our FE-GAN. We first feed the incomplete human parsing, sketch, noise, color, and mask into free-form parsing network to obtain complete synthesized parsing. Then, incomplete image, composed mask, and synthesized parsing are fed into parsing-aware inpainting network for manipulating the image by using the sketch and color.}
    \label{fig:framework}
\end{figure}

We propose a novel method for editing fashion image, allowing users to edit images with a few sketches and sparse color strokes on an interested region. The overview of our FE-GAN is shown in Figure~\ref{fig:framework}. The main components of our FE-GAN include a free-form parsing network and a parsing-aware inpainting network with the multi-scale attention normalization layers. We first discuss the free-form parsing network in Section~\ref{s:fpn}. It can manipulate human parsing guided by free-form sketch and color, and is crucial to help the parsing-aware inpainting network produce convincing interactive results, which is described in Section~\ref{s:pin}. Then, in Section~\ref{s:anl}, we describe the attention normalization layers inserted at multiple scales in the inpainting decoder that can selectively extract effective features and enhance visual quality. Finally, in Section~\ref{s:lof}, we give a detailed description of the learning objective function used in our FE-GAN.

\subsection{Free-form Parsing Network}
\label{s:fpn}
Compared to directly restoring an incomplete image, predicting a parsing map from an incomplete parsing map is more feasible since there are fewer details in the parsing map. Meanwhile, the semantic information in the parsing map can be a guidance for rendering detail textures in each part of an image precisely. To this end, we propose a free-form parsing network to synthesize a complete parsing map when giving an incomplete parsing map and arbitrary sketch and color strokes. 

The architecture of the free-form parsing network is illustrated in the upper left part of Figure~\ref{fig:framework}. It is based on the encoder-decoder architecture like U-net~\cite{ronn2015unet}. The encoder receives five inputs: an incomplete parsing map, a binary sketch that describes the structure of the removed region, a noise sampled from the Gaussian distribution, sparse color strokes and a mask. More details about the input data will be discussed in Section~\ref{s:detail}. It is worth noting that given the same incomplete parsing map and various sketch and color strokes, the free-form parsing network can synthesize different parsing map, which indicates that our parsing generation model is controllable. It is significant for our fashion editing system since different parsing maps guide to render different contents in the edited image.

\subsection{Parsing-aware Inpainting Network}
\label{s:pin}
The architecture of parsing-aware inpainting network is illustrated on the bottom of Figure~\ref{fig:framework}. Inspired by~\cite{Liu2018ImageIF}, we introduce a partial convolution encoder to extract feature from the valid region in incomplete images.
Our proposed partial convolution in partial convolution encoder is a bit different from the original version. Instead of using the mask directly, we utilize the composed mask to make the network focus only on the foreground region. The composed mask can be expressed as:
\begin{equation}
    \mathbf{M}^{\prime}=(\mathbf{1}-\mathbf{M})\odot\mathbf{M}_{\text{f}},
\end{equation}
where $\mathbf{M}^{\prime}$, $\mathbf{M}$ and $\mathbf{M}_\text{f}$ are the composed mask, original mask and foreground mask respectively. $\odot$ denotes element-wise multiply. 
Besides the partial convolution encoder, we introduce a standard convolution encoder to extract semantics feature from the synthesized parsing map. The human parsing map has semantics and location information that will guide the inpainting, since the content in a region with the same semantics should be similar. Given the semantic features, the network can render textures on the particular region more precisely. 
Two encoded feature maps are concatenated together in a channel-wise manner. Then the concatenated feature map undergoes several dilated residual blocks. During the upsampling process, well-designed multi-scale attention normalization layers are introduced to obtain attention maps, which are conditioned on sketch and color strokes. Unlike SC-FEGAN, the learned attention maps are helpful to select more effective feature in the forward activations. We explain the details in the next section.

\subsection{Attention Normalization Layers}
\label{s:anl}
Attention Normalization Layers (ANLs) are similar to SPADE~\cite{park2019spade} to some extent and can be regarded as a variant of conditional normalization. However, instead of inferring an affine transformation from external data directly, ANLs learn an attention map which is used to extract the significant information in the earlier normalized activation. The upper right part of Figure~\ref{fig:framework} illustrates the design of ANLs. The details of ANLs are shown below.

Let $x^i$ denotes the activations of the layer $i$ in the deep neural network. Let $N$ denotes the number of samples in one batch. Let $C^i$ denotes the number of channels of $x^i$. Let $H^i$ and $W^i$ represent the height and width of activation map in layer $i$ respectively. When the activations $x^i$ passing through ANLs, they are first normalized in a channel-wise manner. Then the normalized activations are modulated by the learned attention map and bias. Finally, the modulated activations pass through a rectified linear unit (RELU) and a convolution layer and concatenate with the original normalized activations. The activations value before the final concatenation at position $(n \in N,c \in C^i, h \in H^i, w \in W^i)$ is signed as:
\begin{equation}
    f(\alpha^i_{c,h,w}(\textbf{d})\frac{x^i_{n,c,h,w}-\mu^i_c}{\sigma^i_c}+\beta^i_{c,h,w}(\textbf{d})), 
\end{equation}
where $f(x)$ denotes RELU and convolution operations, $x^i_{n,c,h,w}$ is the activation value at particular position before normalization, $\mu^i_c$ and $\sigma^i_c$ are the mean and standard deviation of activation in channel $c$. As the same of BN~\cite{ioffe2015batch}, we formulate them as:
\begin{equation}
    \mu^i_c = \frac{1}{NH^iW^i}\sum_{n,h,w}x^i_{n,c,h,w}
\end{equation}
\begin{equation}
\sigma^i_c = \sqrt{\frac{1}{NH^iW^i}\sum_{n,h,w}(x^i_{n,c,h,w})^2-(\mu^i_c)^2}
\end{equation}
The $\alpha^i_{c,h,w}(\textbf{d})$ and $\beta^i_{c,h,w}(\textbf{d})$ are learned attention map and bias for modulating the normalization layer, which are conditioned on the external data $\textbf{d}$, namely, the sketch and color strokes and noise in this paper. Our implementations of $\alpha^i_{n,h,w}$ and $\beta^i_{n,h,w}$ are straightforward. The external data is first projected into an embedding space through a convolution layer. Then the bias is produced by another convolution layer, and the attention map is generated by a convolution layer and a sigmoid operation, which limits the range of feature map values between zero and one, and ensures the output to be an attention map.
The effectiveness of ANLs is due to their inherent characteristics. Similar to SPADE \cite{park2019spade}, ANLs can avoid washing away semantic information in activations, since the attention map and bias are spatially-varying. Moreover, the multi-scale ANLs can not only adapt the various scales of activations during upsampling but also extract coarse-to-fine semantic information from external data, which guide the fashion editing more precisely.


\subsection{Learning Objective Function}
\label{s:lof}
Due to the complex textures of the incomplete image and the variety of sketch and color strokes, the training of the free-form parsing network and parsing-aware inpainting network is a challenging task. To address these problems, we apply several losses to make the training easier and more stable in different aspects. Specifically, we apply 
adversarial loss $\mathcal{L}_{\text{adv}}$~\cite{goodfellow2014generative}, 
perceptual loss $\mathcal{L}_{\text{perceptual}}$~\cite{johnson2016perceptual},
style loss $\mathcal{L}_{\text{style}}$~\cite{johnson2016perceptual},
parsing loss $\mathcal{L}_{\text{parsing}}$~\cite{gong2017look},
multi-scale feature loss $\mathcal{L}_{\text{feat}}$~\cite{wang2017pix2pixHD},
and total variation loss $\mathcal{L}_{\text{TV}}$~\cite{johnson2016perceptual} to regularize the training. 
We define a face TV loss to remove the artifacts of the face by using $\mathcal{L}_{\text{TV}}$ on face region.
We define a mask loss by using the L1 norm on the mask area, let $I_{\text{gen}}$ be generated image, let $I_{\text{real}}$ be ground truth,  and let $M$ be the mask, which is computed as:
\begin{equation}
    \mathcal{L}_{\text{mask}}=||I_{\text{gen}} \odot M - I_{\text{real}}
    \odot M ||_1 ,
\end{equation}
we also define a foreground loss to enhance the foreground quality. Let $M_{\text{foreground}}$ be the mask of foreground part, then $\mathcal{L}_{\text{foreground}}$ can be formally computed as
\begin{equation}
    \mathcal{L}_{\text{foreground}}=||I_{\text{gen}} \odot M_{\text{foreground}} - I_{\text{real}}  \odot M_{\text{foreground}}||_1 ,
\end{equation}
similar to $\mathcal{L}_{\text{foreground}}$, we formulate a face loss $\mathcal{L}_{\text{face}} $ to improve the quality of face region. 

The overall objective function $\mathcal{L}_{\text{free-form-parser}}$ for free-form parsing network is formulated as:
\begin{equation}
    \mathcal{L}_{\text{free-form-parser}}=\gamma_1\mathcal{L}_{\text{parsing}}+\gamma_2\mathcal{L}_{\text{feat}}+\gamma_3\mathcal{L}_{\text{adv}} ,
\end{equation}
where hyper-parameters $\gamma_1$, $\gamma_2$ and $\gamma_3$ are weights of each loss.

The overall objective function $\mathcal{L}_{\text{inpainter}}$ for parsing-aware inpainting network written as:
\begin{equation}
	\mathcal{L}_{\text{inpainter}}=
	\lambda_1\mathcal{L}_{\text{mask}}         
	+\lambda_2\mathcal{L}_{\text{foreground}}
	+ \lambda_3\mathcal{L}_{\text{face}}   
	+\lambda_4\mathcal{L}_{\text{faceTV}}
	+\lambda_5\mathcal{L}_{\text{perceptual}}
	+\lambda_6\mathcal{L}_{\text{style}}
	+\lambda_7\mathcal{L}_{\text{adv}} ,
\end{equation}
where hyper-parameters $\lambda_i,(i=1,2,3,4,5,6,7)$ are the weights of each loss.

\begin{figure}[ht]
    \centering
    \includegraphics[width=1.0\hsize \hspace{0.01\hsize}]{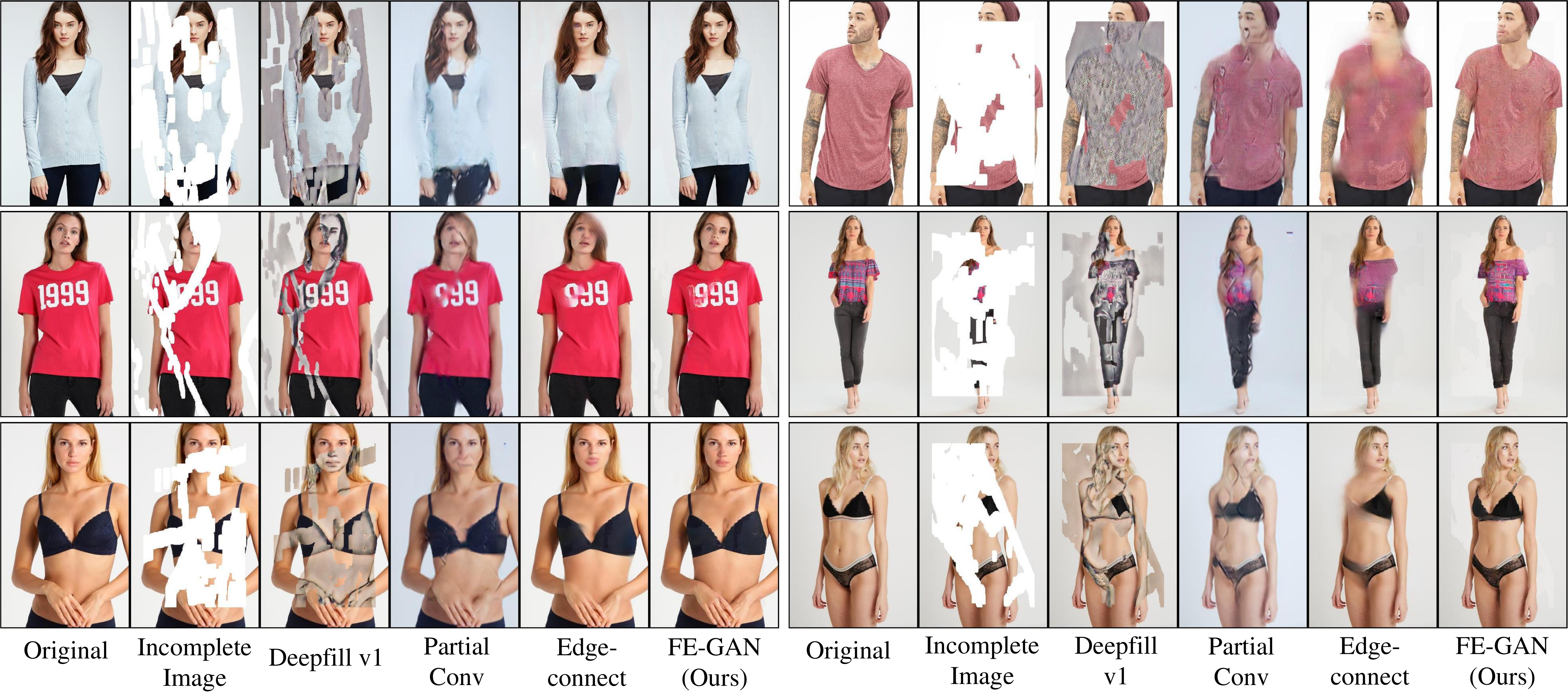} 
    \caption{Qualitative comparisons with Deepfill v1~\cite{Yu2018GenerativeII}, Partial Conv~\cite{Liu2018ImageIF}, and Edge-connect~\cite{Nazeri2019EdgeConnectGI} on DeepFashion~\cite{liu2016deepfashion}, MPV~\cite{dong2019towards}, and FashionE, respectively.}
    \vspace{-2mm}
    \label{fig:visual}
    \vspace{-2mm}
\end{figure}
\section{Experiments}

\subsection{Datasets and Metrics}






We conduct our experiments on \textbf{DeepFashion}~\cite{liu2016deepfashion} from Fashion Image Synthesis track. It contains 38,237 images which are split into a train set and a test set, 29,958 and 8,279 images respectively. \textbf{MPV}~\cite{dong2019towards} contains 35,687 images which are split into a train set and a test set, 29,469 and 6,218 samples. For better contributing to the fashion editing community, we collected a new fashion dataset, named \textbf{FashionE}. It contains 7,559 images with the size of $320 \times 512$. In our experiment, we split it into a train set of 6,106 images and a test set of 1,453 images. The dataset will be released upon the publication of this work. The size of the image is $320 \times 512$ across all datasets.

We utilize the \textbf{Irregular Mask Dataset} provided by~\cite{Liu2018ImageIF} in our experiments. The original dataset contains 55,116 masks for training and 24,866 masks for testing. We randomly select 12,000 images, splitting it into one train set of 9,600 masks and one test set of 2,400 masks. To mimic the free-form color stroke, we utilize one irregular mask dataset from~\cite{Iskakov2018SemiparametricII} as \textbf{Irregular Strokes Dataset}. The mask region stands for stroke in our experiment. In our experiment, we split it into a train set of 50,000 masks and a test set of 10,000 masks. In our experiment, all the masks are resized to $320 \times 512$.

\textbf{Metrics}.
We evaluate our proposed method, as well as compared approaches on three metrics, PSNR (Peak Signal Noise Ratio), SSIM (Structural Similarity index)~\cite{wang2004image}, and FID (Fréchet Inception Distance)~\cite{Heusel2017GANsTB}. We apply the Amazon Mechanical Turk (AMT) for evaluating the qualitative results.

\subsection{Implementation Details}
\label{s:detail}

\textbf{Training Procedure}. 
The training procedure is two-stage. The first stage is to train free-form parsing network. We use $\gamma_1$ = 10, $\gamma_2$ = 10, $\gamma_3$ = 1 in the loss function. The second stage is to train parsing-aware inpainting network. We use $\lambda_1$ = 5.0, $\lambda_2$ = 50, $\lambda_3$ = 1.0, $\lambda_4$ = 0.1, $\lambda_5$ = 0.05, $\lambda_6$ = 200, $\lambda_7$ = 0.001 in the loss function. For both training stages, we use Adam~\cite{kingma2014adam} optimizer with $\beta_1$ = 0.5 and $\beta_2$ = 0.999 and learning rate is 0.0002. The batch sizes of stage 1 is 20, and stage 2 is 8. In each training cycle, we train one step for the generator and one step for the discriminator. All the experiments are conducted on 4 Nvidia 1080 Ti GPUs. 

\textbf{Sketch \& Color Domain}. 
The way of extracting sketch and color domain from images is similar to SC-FEGAN. Instead of using HED~\cite{Xie2015HolisticallyNestedED}, we generated sketches by Canny Edge Detector~\cite{Canny1986ACA}. Relying on the result of human parsing, we use the median color of each segmented area to represent the color of that area. More details are presented in the supplementary material.

\textbf{Discriminators}.
The discriminator, used in free-form parsing network, has a similar structure as the multi-scale discriminator in Pixel2PixelHD~\cite{wang2017pix2pixHD}, which has two PatchGAN discriminators. The discriminator, used in parsing-aware inpainting network, has a similar structure as inpainting discriminator in Edge-connect~\cite{Nazeri2019EdgeConnectGI}, with five convolutions and spectral norm blocks.

\textbf{Compared Approaches}.
To make a comprehensive evaluation of our proposed method, we conduct three comparison experiments based on the recent state of the art approaches at image inpainting~\cite{Yu2018GenerativeII,Liu2018ImageIF,Nazeri2019EdgeConnectGI}. It comprises of an edge generator and an image completion module. The re-implementations followed the source codes provided by authors. To make a fair comparison, all inputs consist of incomplete images, masks, sketch, color domain, and noise across all comparison experiments.

\begin{figure}[t]
    \centering
    \includegraphics[width=1.0\hsize \hspace{0.01\hsize}]{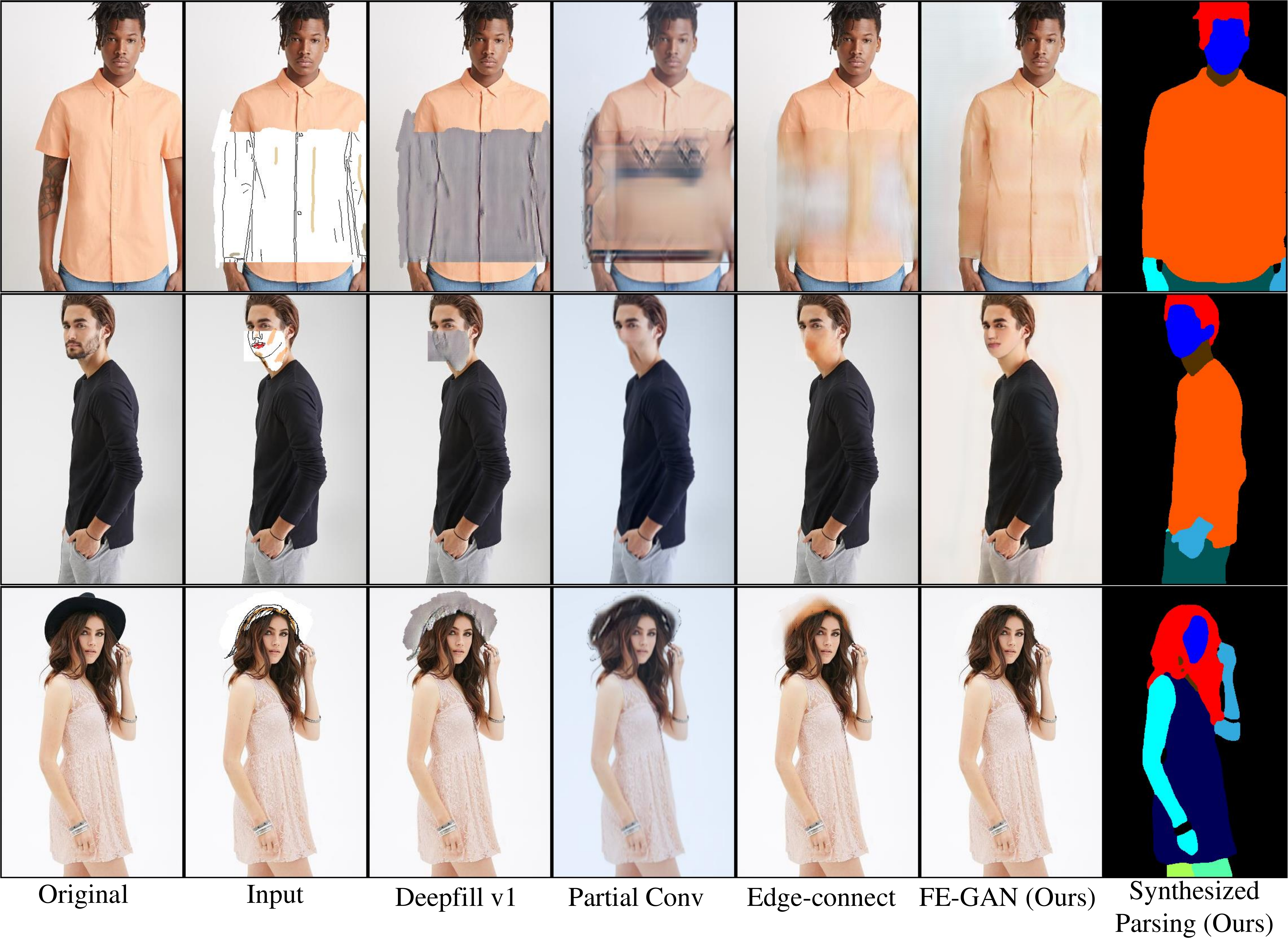}
    \vspace{-4mm}
    \caption{Some interactive comparisons with Deepfill v1~\cite{Yu2018GenerativeII}, Partial Conv~\cite{Liu2018ImageIF}, and Edge-connect~\cite{Nazeri2019EdgeConnectGI} on DeepFashion~\cite{liu2016deepfashion}, MPV~\cite{dong2019towards}, and FashionE, respectively.}
    \vspace{-2mm}
    \label{fig:interactive}
\end{figure}

\begin{table}[t]
\centering
 \caption{Quantitative comparisons on DeepFashion~\cite{liu2016deepfashion}, MPV~\cite{dong2019towards}, and FashionE datasets.}
\resizebox{\columnwidth}{!}{
\begin{tabular}{l ccc | ccc | ccc}
	\toprule
    & \multicolumn{3}{c}{DeepFashion~\cite{liu2016deepfashion}} &  
    \multicolumn{3}{c}{MPV~\cite{dong2019towards}} &
    \multicolumn{3}{c}{FashionE} \\
     \cmidrule{2-10} 
        Model  & PSNR & SSIM & FID     & PSNR & SSIM & FID    & PSNR & SSIM & FID \\
    \midrule
     	Deepfill v1~\cite{Yu2018GenerativeII}   & 16.885 & 0.781 & 60.994      & 18.450 & 0.808 & 58.742      & 19.170 & 0.814 & 56.738 \\
       	Partial Conv~\cite{Liu2018ImageIF}       & 19.103  & 0.827  & 17.728      & 20.408  & 0.850  & 22.751       & 20.635  & 0.848  & 20.148  \\
       	Edge-connect~\cite{Nazeri2019EdgeConnectGI}       & 26.236  & 0.901  & 12.633     & 27.557 & 0.924  & 7.888     & 29.154  & 0.926  & 5.182 \\
    \midrule
        FE-GAN (Ours)     & \textbf{29.552} & \textbf{0.928} & \textbf{3.700}  & \textbf{30.602} & \textbf{0.944} & \textbf{3.796}  & \textbf{30.974}  & \textbf{0.938} & \textbf{3.246}       \\
    \bottomrule
\end{tabular}
}
\label{tab:score}
\vspace{-4mm}
\end{table}


\subsection{Quantitative Results}
PSNR computes the peak signal-to-noise ratio between images. SSIM measures the similarity between two images. Higher value of PSNR and SSIM mean better results. FID is tended to replace Inception Score as one of the most significant metrics measuring the quality of generated images. It computes the Fréchet distance between two multivariate Gaussians, the smaller the better. As mentioned in~\cite{wang2004image}, there is no good numerical metric in image inpainting. Furthermore, our focus is even beyond the regular inpainting. We can observe from Table~\ref{tab:score}, our FE-GAN achieves the best PSNR, SSIM, and FID scores and outperforms all other methods among three datasets.


\subsection{Qualitative Results}
Beyond numerical evaluation, we present visual comparisons for image completion task among three datasets and four methods, shown in Figure~\ref{fig:visual}. Three rows, from top to bottom, are results from DeepFashion, MPV, and FashionE. The interactive results for those methods are shown in Figure~\ref{fig:interactive}. The last column of the Figure~\ref{fig:interactive}, are the results of the free-form parsing network. We can observe that the free-form parsing network can obtain promising parsing results by manipulating the sketch and color. Thanks to the multi-scale attention normalization layers and the synthesized parsing result from the free-form parsing network, our FE-GAN outperforms all other baselines on visual comparisons.


\subsection{Human Evaluation}
To further demonstrate the robustness of our proposed FE-GAN, we conduct the human evaluation deployed on the Amazon Mechanical Turk platform on the DeepFashion~\cite{liu2016deepfashion}, MPV~\cite{dong2019towards}, and FashionE. In each test, we provide two images, one from compared methods, the other from our proposed method. Workers are asked to choose the more realistic image out of two. During the evaluation, $K$ images from each dataset are chosen, and $n$ workers will only evaluate these $K$ images. In our case, $K=100$ and $n=10$. We can observe from Table~\ref{tab:human}, our proposed method has a superb performance over the other baselines. This confirms the effectiveness of our FE-GAN comprised of a free-form parsing network and a parsing-aware network, which generates more realistic fashion images.


\begin{table}[h]
\vspace{-2mm}
\centering
\caption{Human evaluation results of pairwise comparison with other methods.}
\begin{tabular}{l c c c}
\toprule
Comparison Method Pair & DeepFashion~\cite{liu2016deepfashion}          & MPV~\cite{dong2019towards}                  & FashionE             \\
\midrule
Ours  \emph{vs} Deepfill v1~\cite{Yu2018GenerativeII}       & \textbf{0.849}  \emph{vs} 0.151 & \textbf{0.845}  \emph{vs} 0.155 & \textbf{0.857}  \emph{vs} 0.143 \\ 
Ours  \emph{vs} Partial Conv~\cite{Liu2018ImageIF}      & \textbf{0.917}  \emph{vs} 0.083 & \textbf{0.864}  \emph{vs} 0.136 & \textbf{0.799}  \emph{vs} 0.201 \\
Ours  \emph{vs} Edge-connect~\cite{Nazeri2019EdgeConnectGI}      & \textbf{0.790}  \emph{vs} 0.210 & \textbf{0.691}  \emph{vs} 0.309 & \textbf{0.656}  \emph{vs} 0.344 \\ 
 \bottomrule
\end{tabular}
\label{tab:human}
\end{table}

\section{Ablation Study}
To evaluate the impact of the proposed component of our FE-GAN, we conduct an ablation study on FashionE with using the model of 20 epochs. As shown in Table~3 and Figure~\ref{fig:ablation}, we report the results of the different versions of our FE-GAN. We first compare the results using attention normalization to the results without using it. We can learn that incorporating the attention normalization layers into the decoder of the inpainting module significantly improves the performance of image completion. We then verify the effectiveness of the proposed free-from parsing network. From Table~3 and Figure~\ref{fig:ablation}, we observe that the performance drops dramatically without using parsing, which can depict the human layouts for guiding image manipulation with higher-level structure constraints. The results report that the main improved performance achieved by the attention normalization and human parsing. We also explore the impact of our designed objective function that each of the losses can substantially improve the results.

\begin{minipage}{\textwidth}
  \begin{minipage}[b]{0.49\textwidth}
    \centering
    \begin{tabular}{lccc}
        \toprule
        Method & PSRN   & SSIM   & FID     \\
        \midrule
        Full    & \textbf{30.035} & \textbf{0.932} & \textbf{4.092}  \\ 
        w/o attention norm        & 29.185 & 0.920 & 5.191 \\ 
        w/o parsing        & 29.109 & 0.923 &  5.355 \\ 
        w/o $\mathcal{L}_{\text{mask}}$        & 28.813 & 0.921 & 4.773  \\ 
        w/o $\mathcal{L}_{\text{foreground}}$        & 29.848 & 0.927 & 5.030  \\ 
        \bottomrule
      \end{tabular}
      \label{tab:ablation}
      \captionof{table}{Ablation studies on FashionE.}
    \end{minipage}
    \hfill
  \begin{minipage}[b]{0.49\textwidth}
    \centering
  \includegraphics[width=0.8\hsize \hspace{0.01\hsize}]{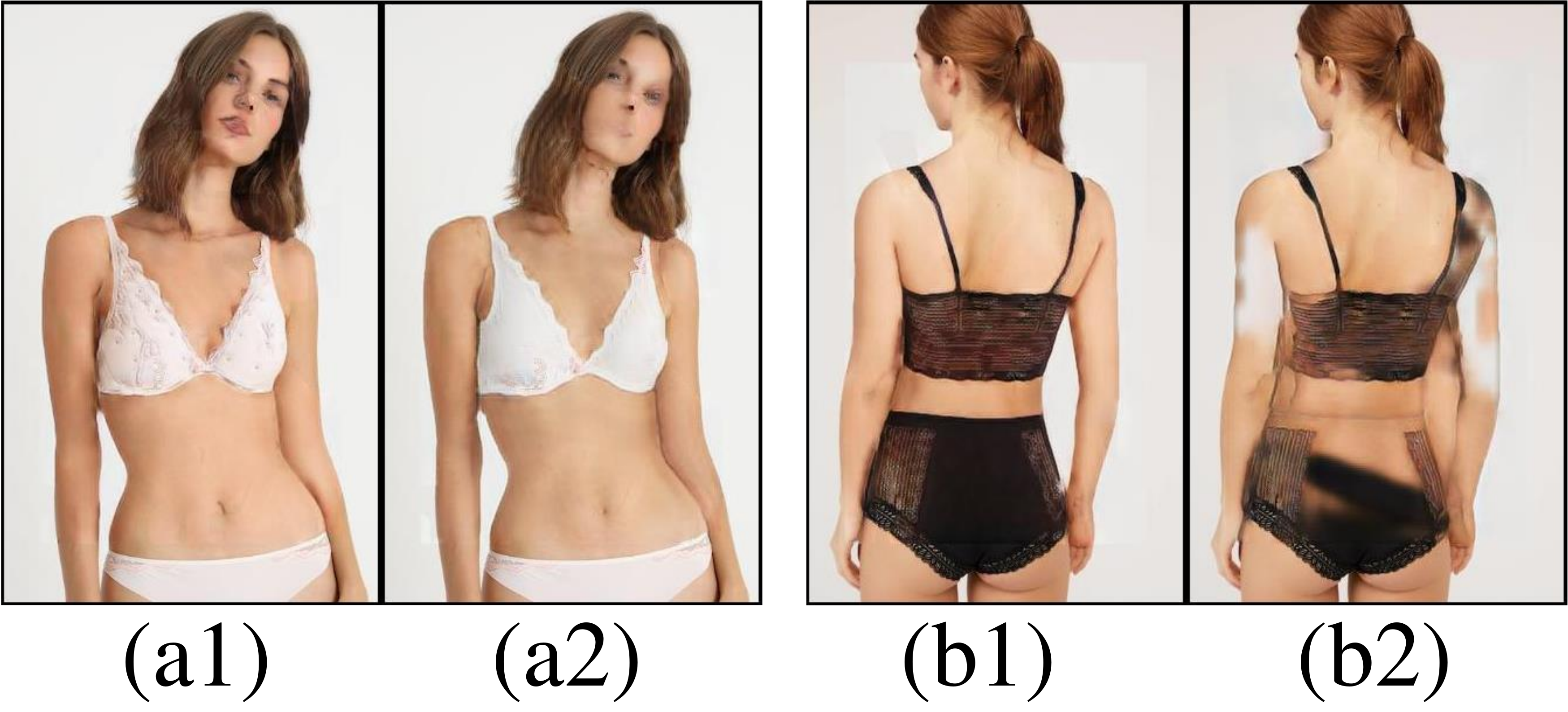}
    \label{fig:ablation}
    \captionof{figure}{Ablation studies on FashionE. (a1)(b1): Ours(Full); (a2): w/o attention norm; (b2): w/o parsing.}
  \end{minipage}
\end{minipage}

\section{Conclusions}
We propose a novel Fashion Editing Generative Adversarial Network (FE-GAN), which enables users to manipulate the fashion image with an arbitrary sketch and a few sparse color strokes. The FE-GAN incorporates a free-form parsing network to predict the complete human parsing map to guide fashion image manipulation. Moreover, we develop a foreground-based partial convolutional encoder and design an attention normalization layer which used in the multiple scales layers of the decoder for the inpainting network. The experiments on fashion datasets demonstrate that our FE-GAN outperforms the state-of-the-art methods and achieves high-quality performance with convincing details. 

\bibliographystyle{plain}
\bibliography{neurips_2019} 

\clearpage
\section*{Appendix}
\begin{figure}[h]
\vspace{-6mm}
    \centering
    \includegraphics[width=1.0\hsize \hspace{0.01\hsize}]{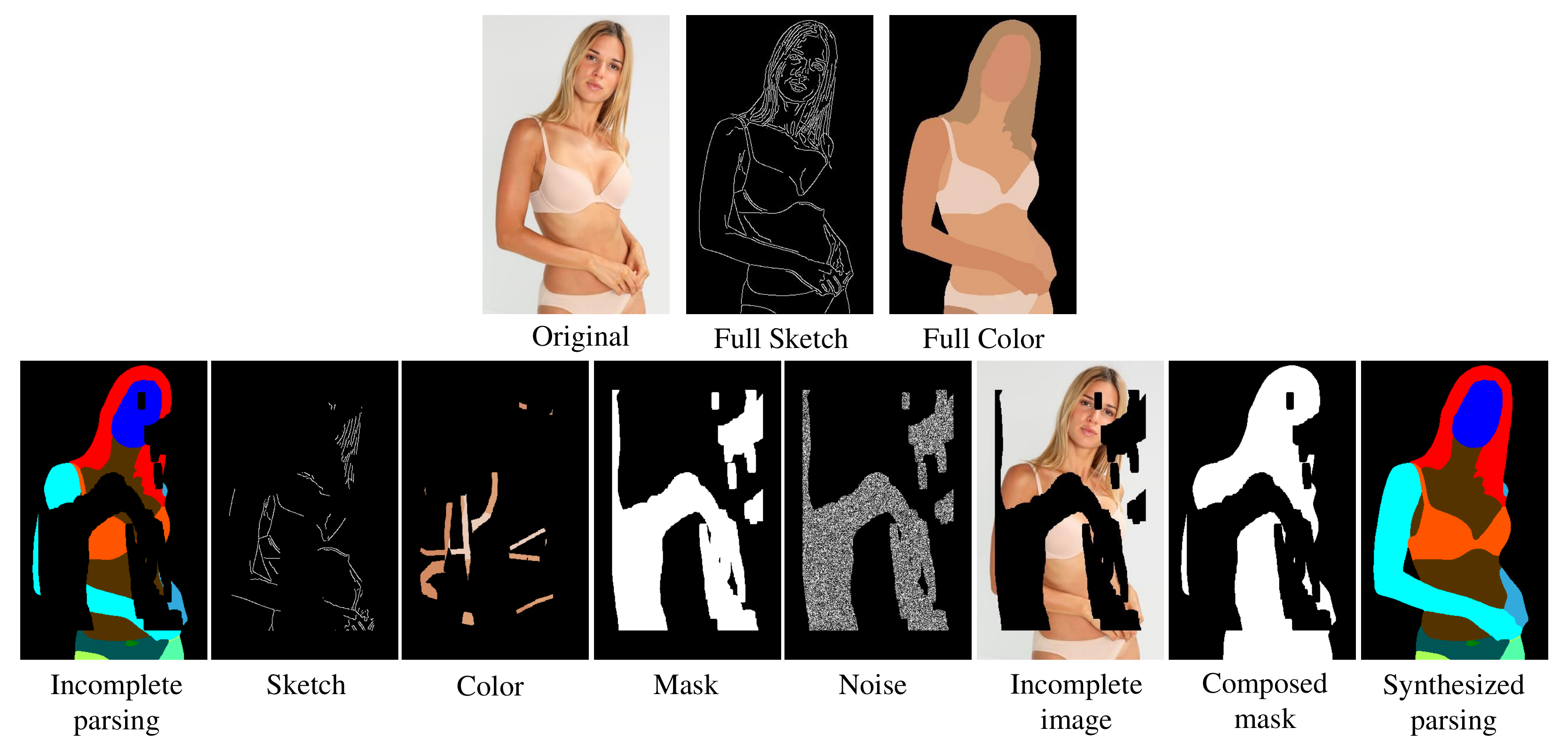} 
    \caption{Example of model inputs shown in the second row. The inputs of the free-form parsing network consist of incomplete parsing, sketch, color, mask, and noise; the inputs of parsing-aware inpainting network contain incomplete image, composed mask and synthesized parsing. The inputs of attention normalization layers are a sketch, color, and noise. We first generate the sketches by using Canny~\cite{Canny1986ACA} shown in the second column of the first row. Then, we use a human parser~\cite{gong2017look} to extract the median color of each part of the person, shown in the last column of the first row. 
    }
    \label{fig:input}
    \vspace{-2mm}
\end{figure}

\begin{figure}[h]
    \centering
    \includegraphics[width=0.9\hsize \hspace{0.01\hsize}]{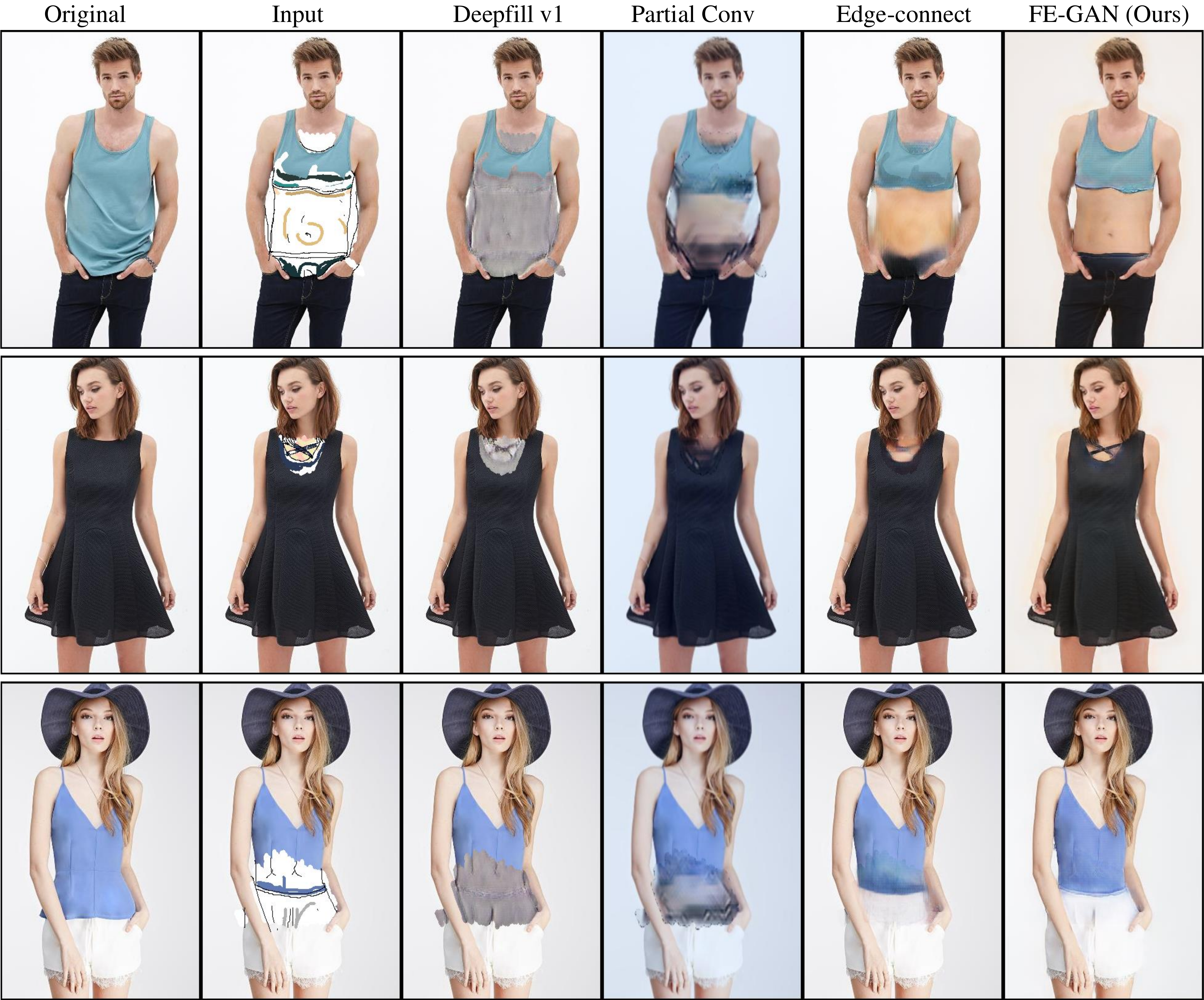} 
    \caption{Some interactive comparisons of Deepfill v1~\cite{Yu2018GenerativeII}, Partial Conv~\cite{Liu2018ImageIF}, Edge-connect~\cite{Nazeri2019EdgeConnectGI}, and FE-GAN (Ours). The results of our FE-GAN are shown in the last column. Zoom in for details.}
    \label{fig:ite3}
   \vspace{-6mm}
\end{figure}

\begin{figure}[ht]
    \centering
    \includegraphics[width=0.99\hsize
    \hspace{0.01\hsize}]{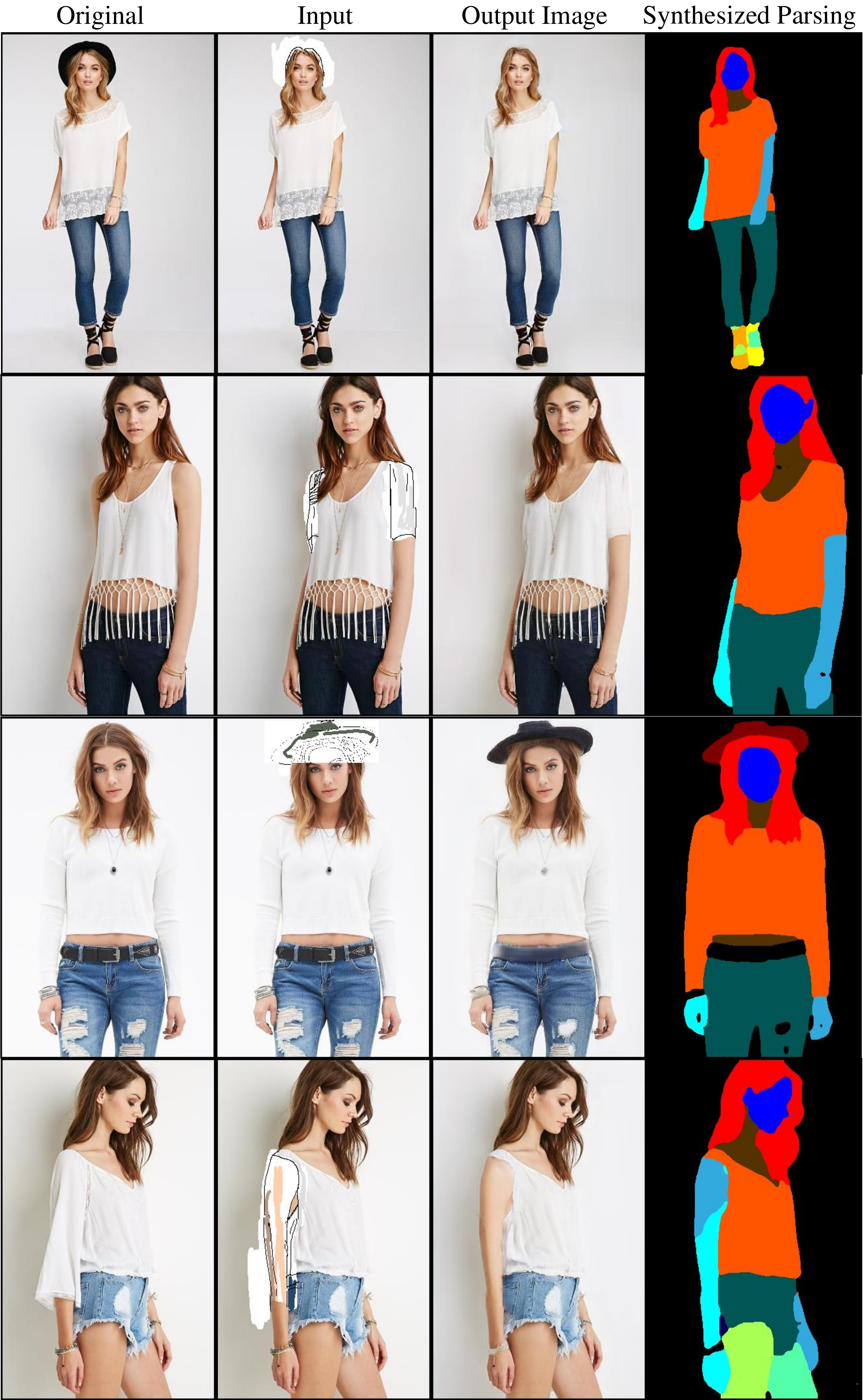} 
    \caption{Some interactive results of our FE-GAN, shown in the third column. The input contains free-form mask, sketch, and sparse color strokes. The results of our free-form parsing network shown in the last column. Zoom in for details.}
    \label{fig:ite1}
\end{figure}

\begin{figure}[ht]
    \centering
    \includegraphics[width=0.99\hsize
    \hspace{0.01\hsize}]{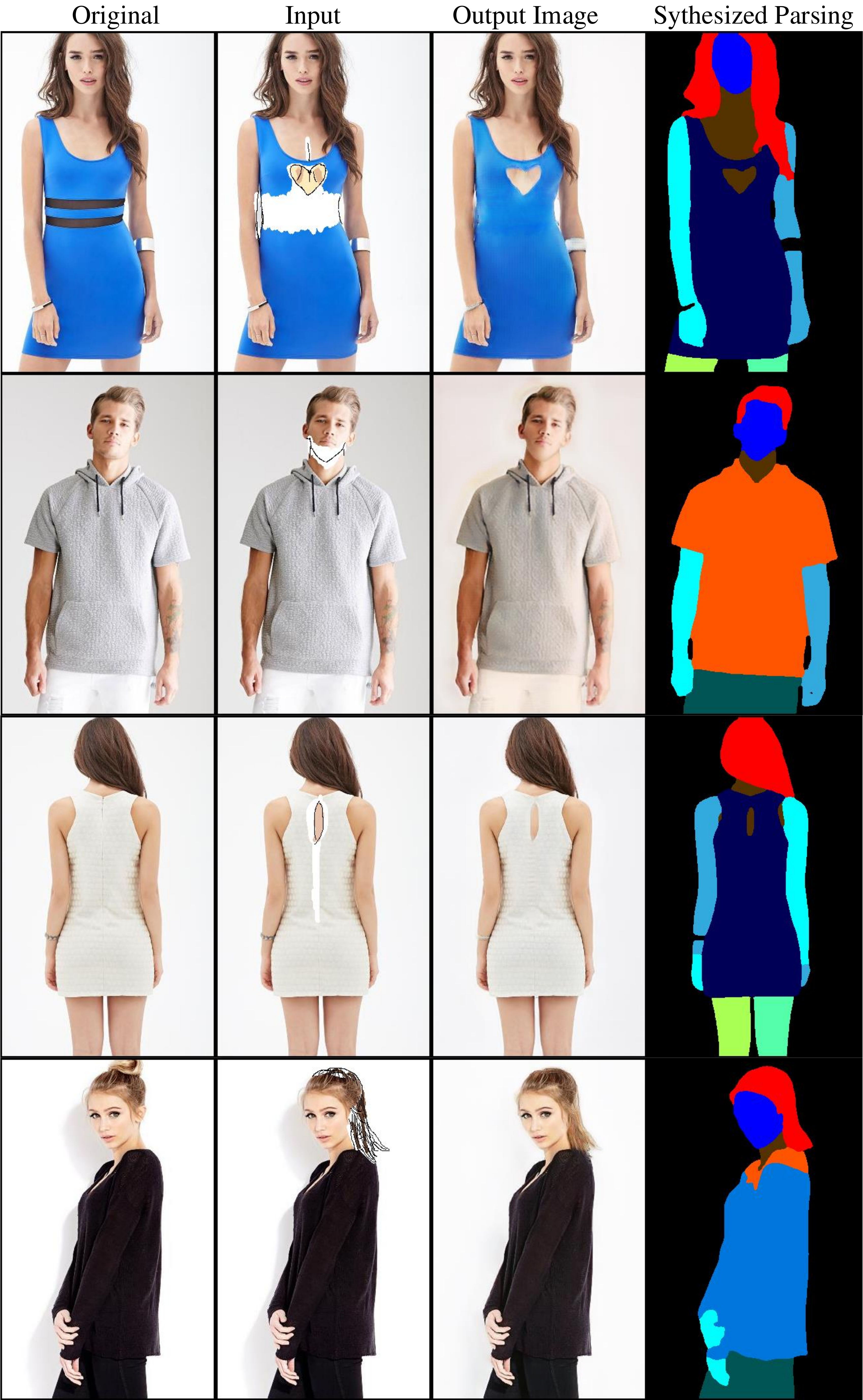} 
    \caption{Some interactive results of our FE-GAN, shown in the third column. The input contains free-form mask, sketch, and sparse color strokes. The results of our free-form parsing network shown in the last column. Zoom in for details.}
    \label{fig:ite2}
\end{figure}

\begin{figure}[ht]
    \centering
    \includegraphics[width=1.0\hsize \hspace{0.01\hsize}]{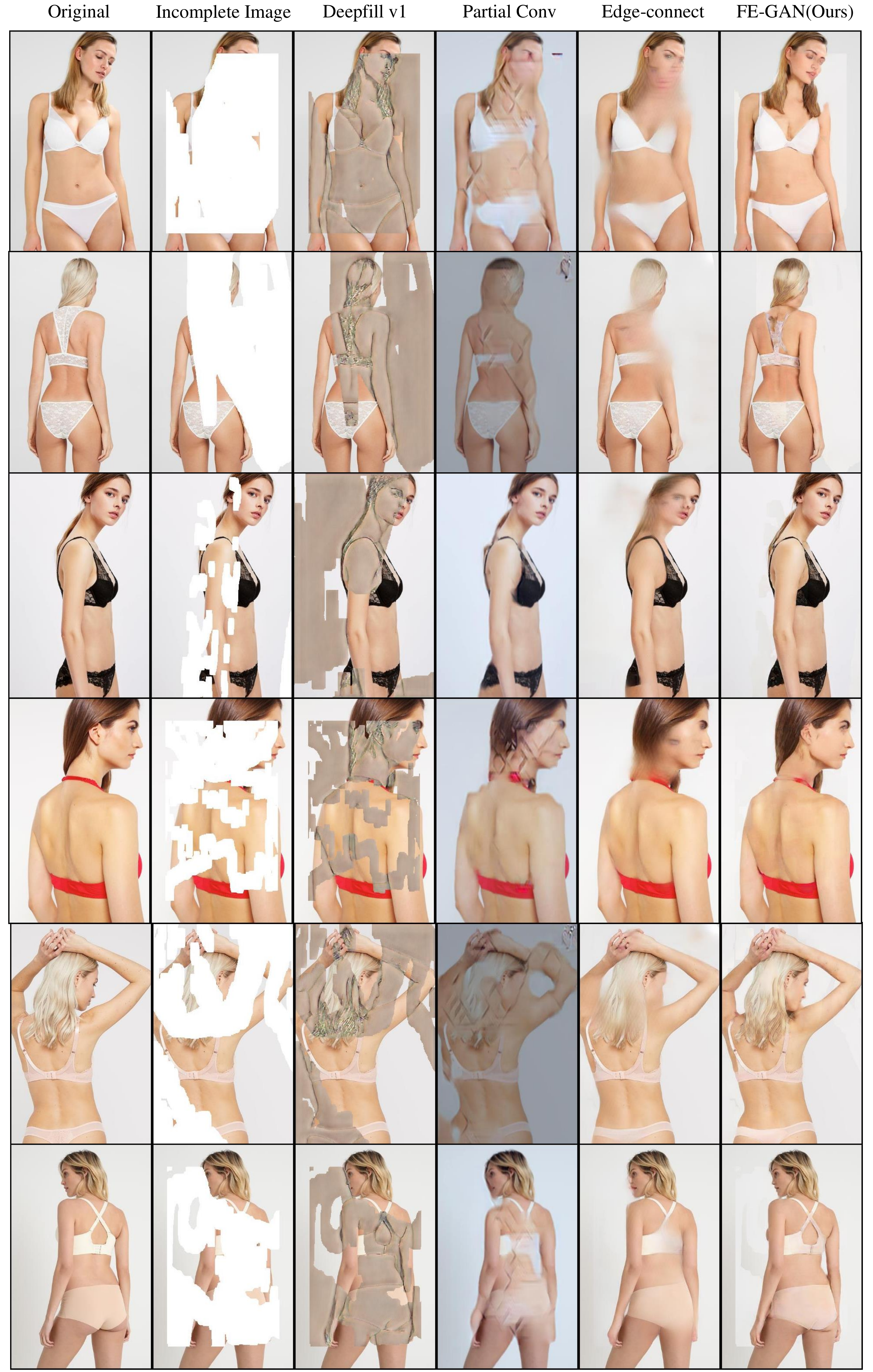} 
    \caption{Qualitative comparisons between Deepfill v1~\cite{Yu2018GenerativeII}, Partial Conv~\cite{Liu2018ImageIF}, Edge-connect~\cite{Nazeri2019EdgeConnectGI}, and FE-GAN on FashionE.}
    \label{fig:bra1}
\end{figure}

\begin{figure}[ht]
    \centering
    \includegraphics[width=1.0\hsize \hspace{0.01\hsize}]{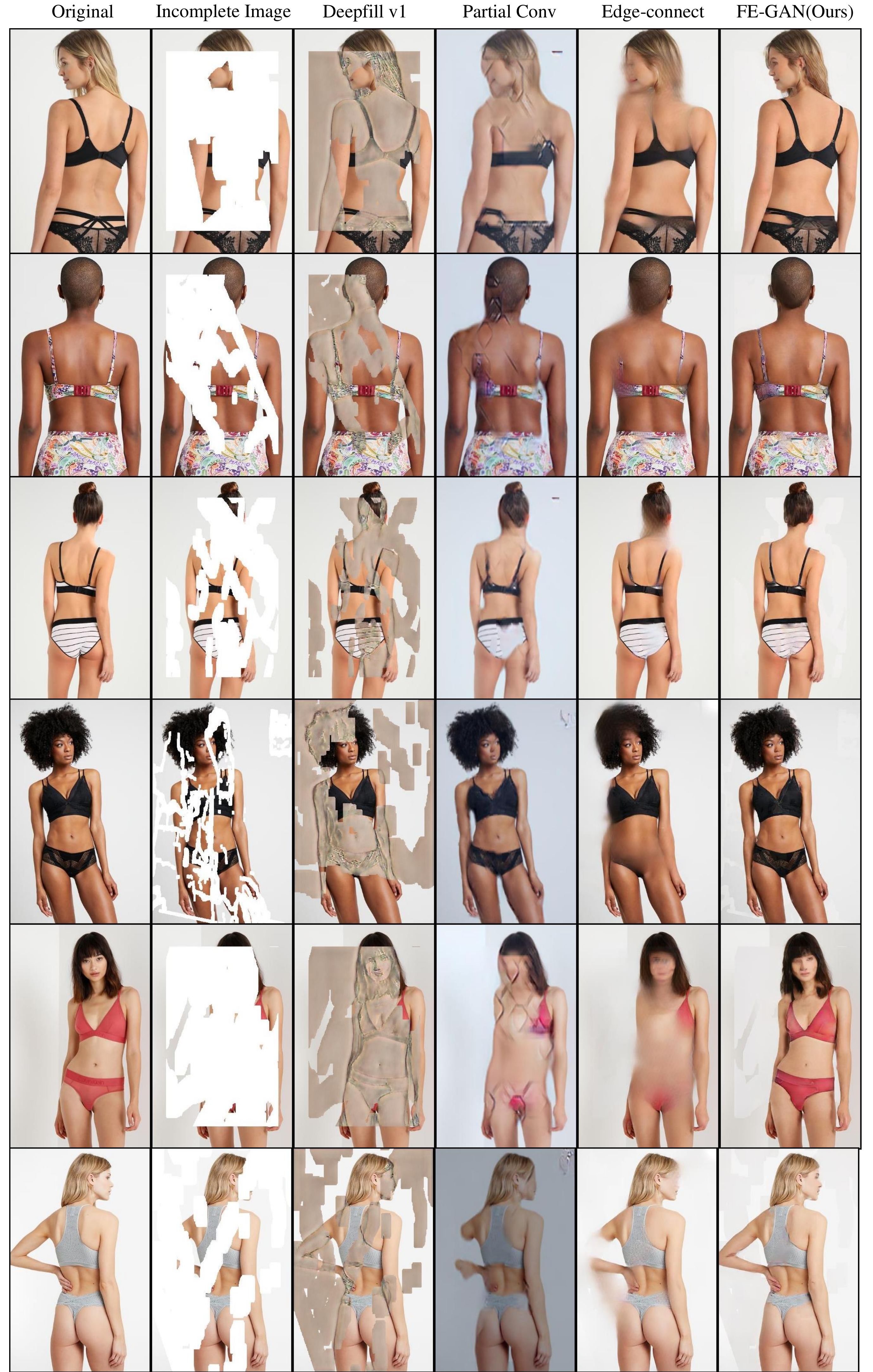} 
    \caption{Qualitative comparisons between Deepfill v1~\cite{Yu2018GenerativeII}, Partial Conv~\cite{Liu2018ImageIF}, Edge-connect~\cite{Nazeri2019EdgeConnectGI}, and FE-GAN on FashionE.}
    \label{fig:bra2}
\end{figure}

\begin{figure}[ht]
    \centering
    \includegraphics[width=1.0\hsize \hspace{0.01\hsize}]{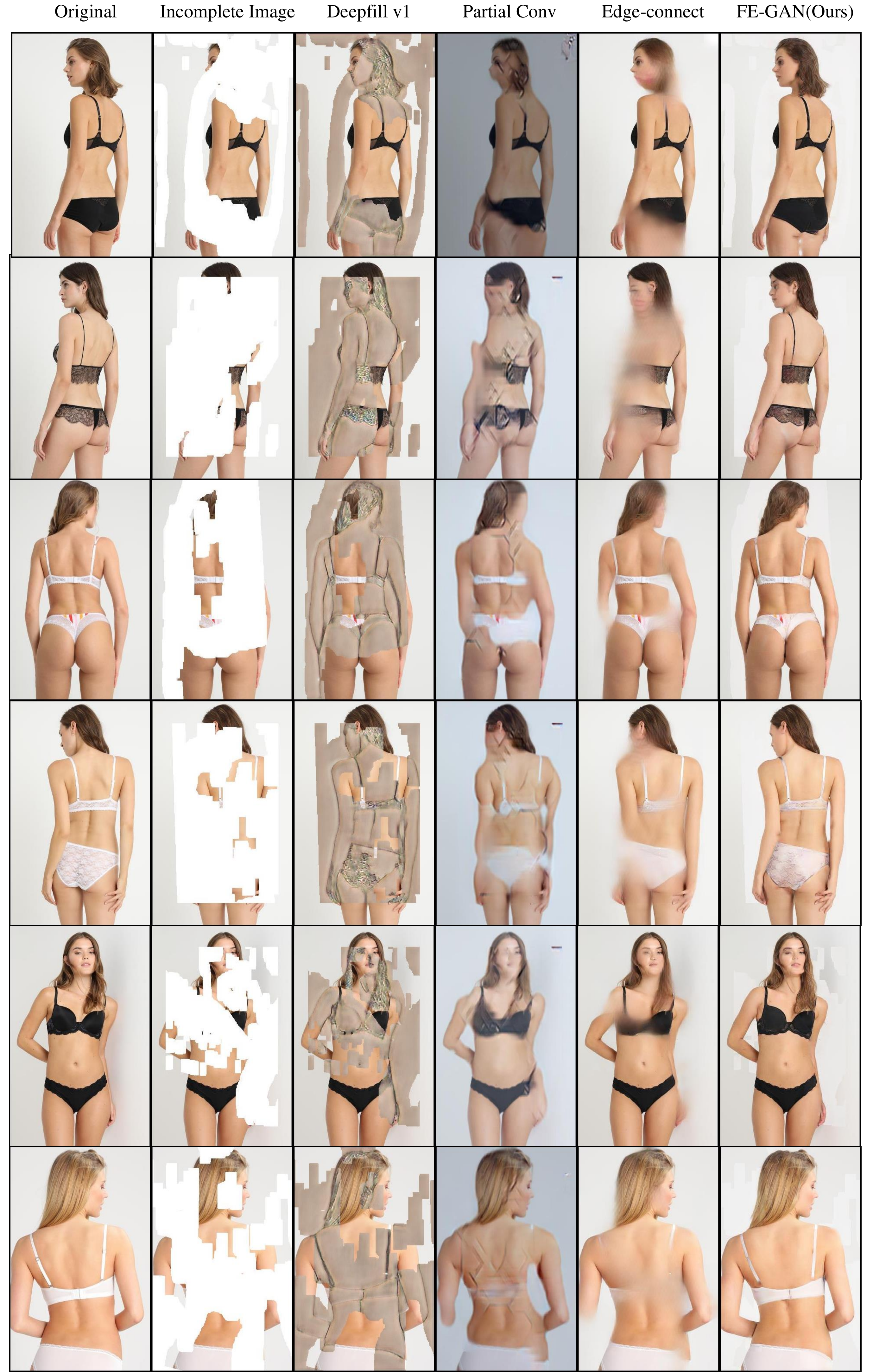} 
    \caption{Qualitative comparisons between Deepfill v1~\cite{Yu2018GenerativeII}, Partial Conv~\cite{Liu2018ImageIF}, Edge-connect~\cite{Nazeri2019EdgeConnectGI}, and FE-GAN on FashionE.}
    \label{fig:bra3}
\end{figure}

\begin{figure}[ht]
    \centering
    \includegraphics[width=1.0\hsize \hspace{0.01\hsize}]{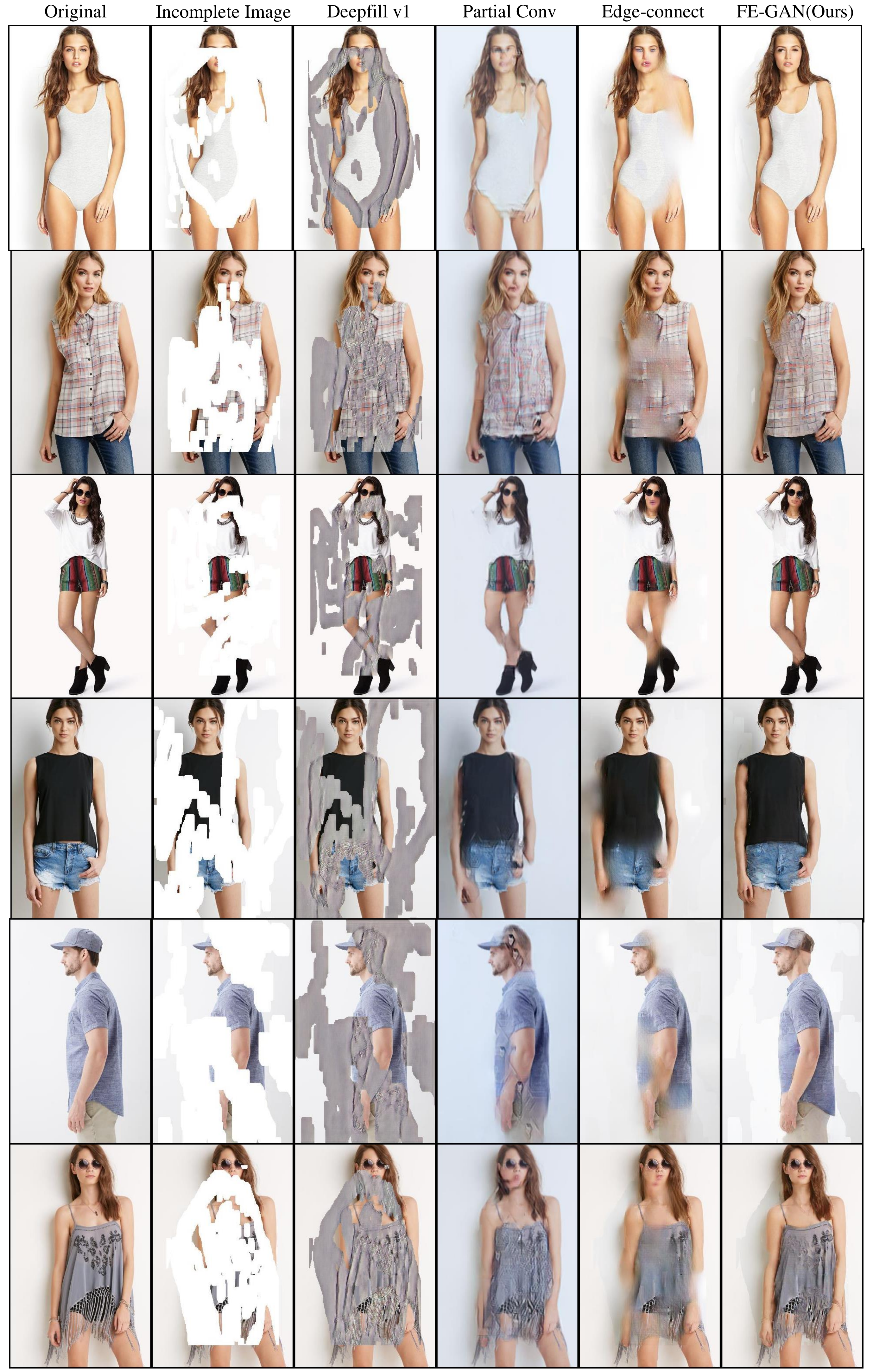} 
    \caption{Qualitative comparisons between Deepfill v1~\cite{Yu2018GenerativeII}, Partial Conv~\cite{Liu2018ImageIF}, Edge-connect~\cite{Nazeri2019EdgeConnectGI}, and FE-GAN on DeepFashion~\cite{liu2016deepfashion}}
    \label{fig:deep-fashion1}
\end{figure}

\begin{figure}[ht]
    \centering
    \includegraphics[width=1.0\hsize \hspace{0.01\hsize}]{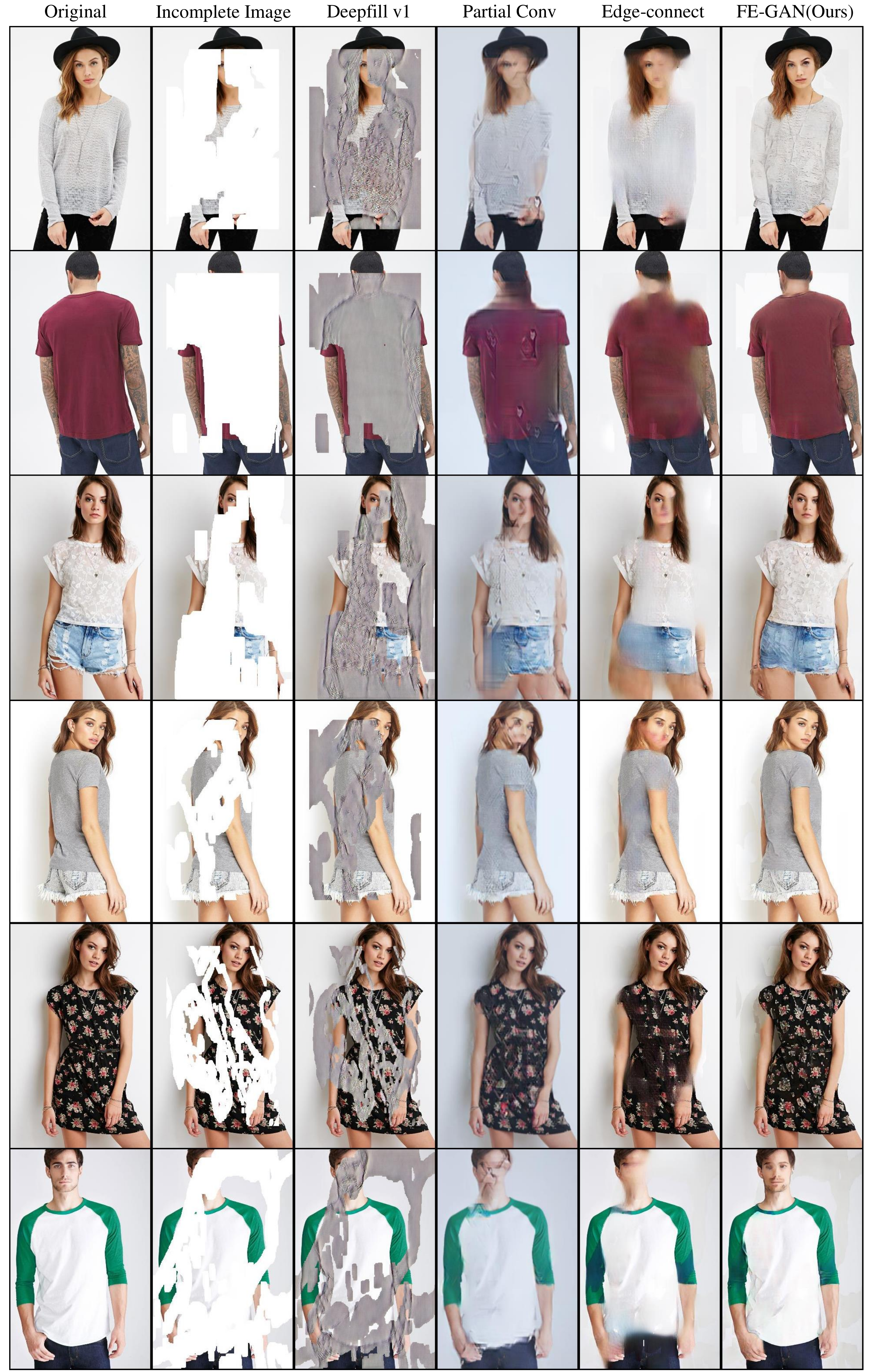} 
    \caption{Qualitative comparisons between Deepfill v1~\cite{Yu2018GenerativeII}, Partial Conv~\cite{Liu2018ImageIF}, Edge-connect~\cite{Nazeri2019EdgeConnectGI}, and FE-GAN on DeepFashion~\cite{liu2016deepfashion}}
    \label{fig:deep-fashion2}
\end{figure}

\begin{figure}[ht]
    \centering
    \includegraphics[width=1.0\hsize \hspace{0.01\hsize}]{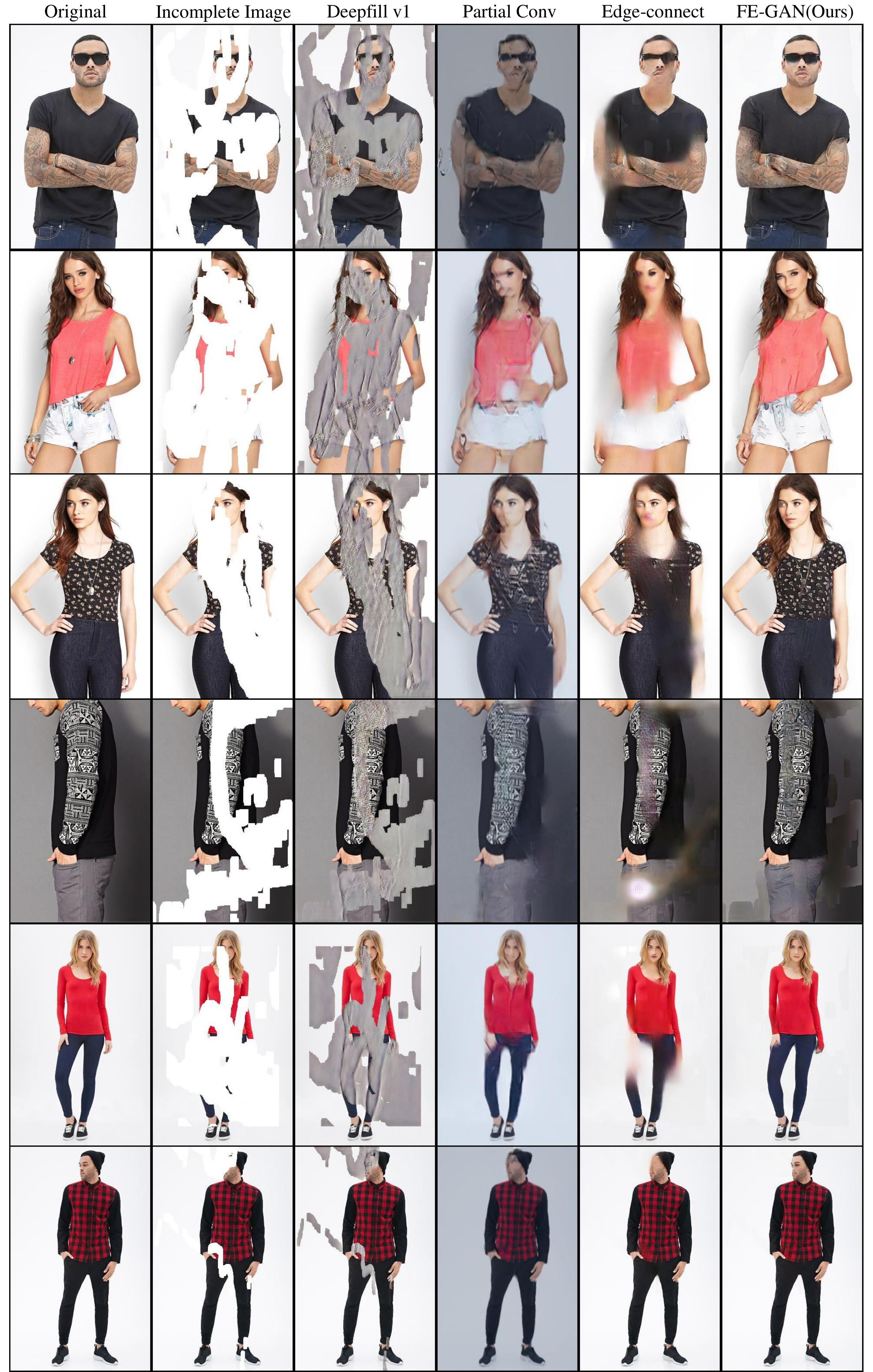} 
    \caption{Qualitative comparisons between Deepfill v1~\cite{Yu2018GenerativeII}, Partial Conv~\cite{Liu2018ImageIF}, Edge-connect~\cite{Nazeri2019EdgeConnectGI}, and FE-GAN on DeepFashion~\cite{liu2016deepfashion}}
    \label{fig:deep-fashion3}
\end{figure}

\begin{figure}[ht]
    \centering
    \includegraphics[width=1.0\hsize \hspace{0.01\hsize}]{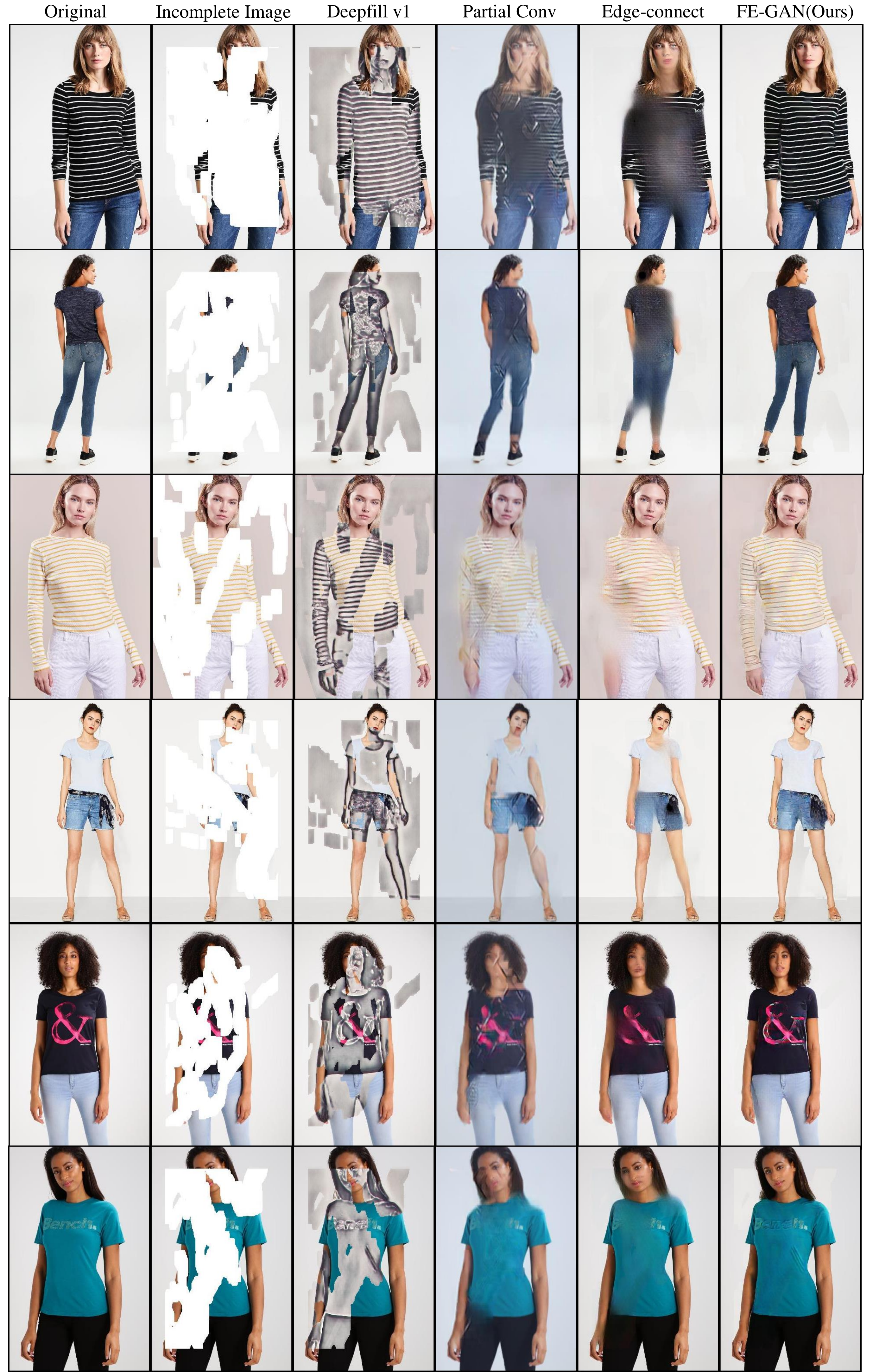} 
    \caption{Qualitative comparisons between Deepfill v1~\cite{Yu2018GenerativeII}, Partial Conv~\cite{Liu2018ImageIF}, Edge-connect~\cite{Nazeri2019EdgeConnectGI}, and FE-GAN on MPV~\cite{dong2019towards}.}
    \label{fig:mpv}
\end{figure}

\begin{figure}[ht]
    \centering
    \includegraphics[width=1.0\hsize \hspace{0.01\hsize}]{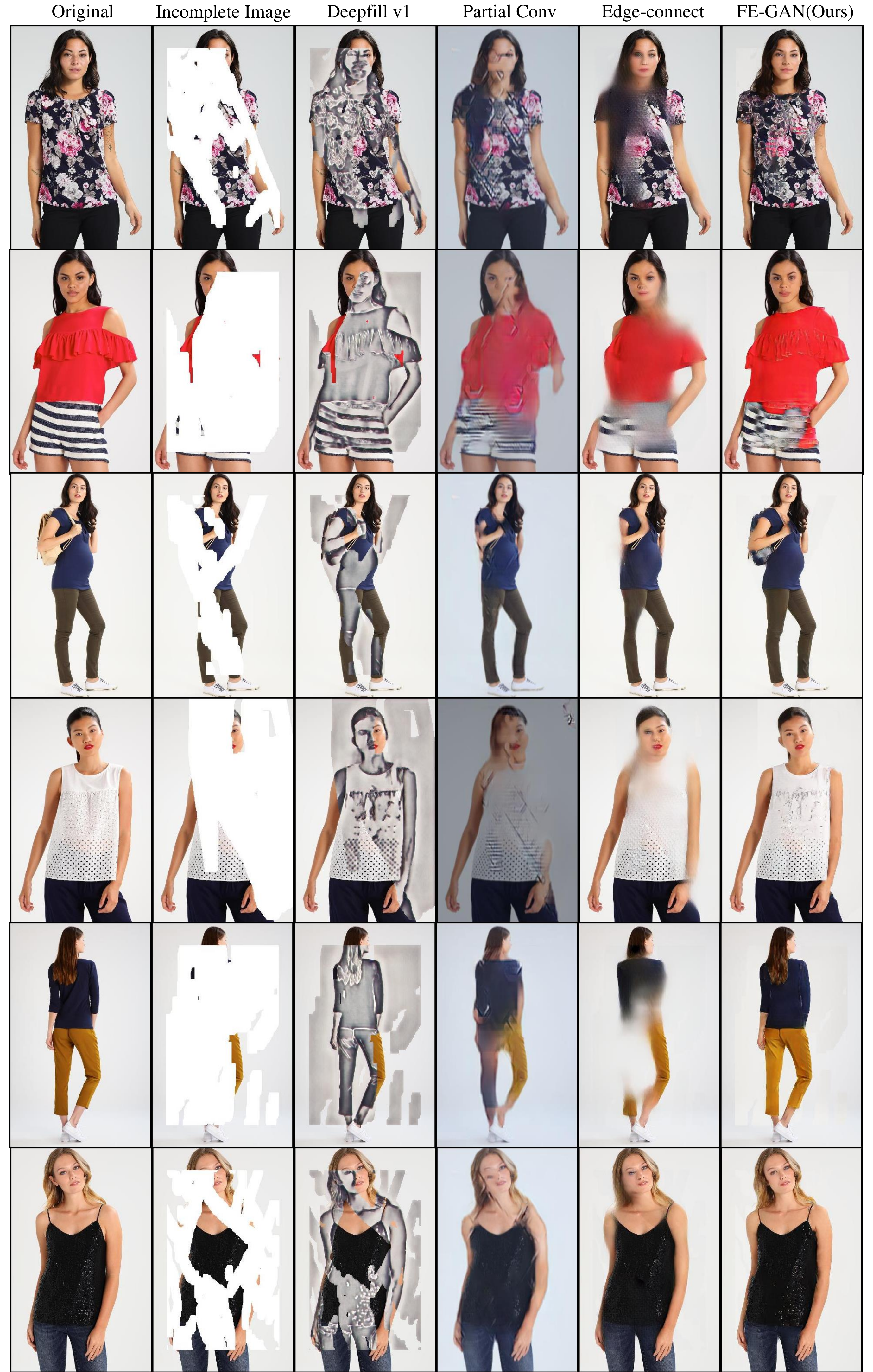} 
    \caption{Qualitative comparisons between Deepfill v1~\cite{Yu2018GenerativeII}, Partial Conv~\cite{Liu2018ImageIF}, Edge-connect~\cite{Nazeri2019EdgeConnectGI}, and FE-GAN on MPV~\cite{dong2019towards}.}
    \label{fig:mpv}
\end{figure}

\begin{figure}[ht]
    \centering
    \includegraphics[width=1.0\hsize \hspace{0.01\hsize}]{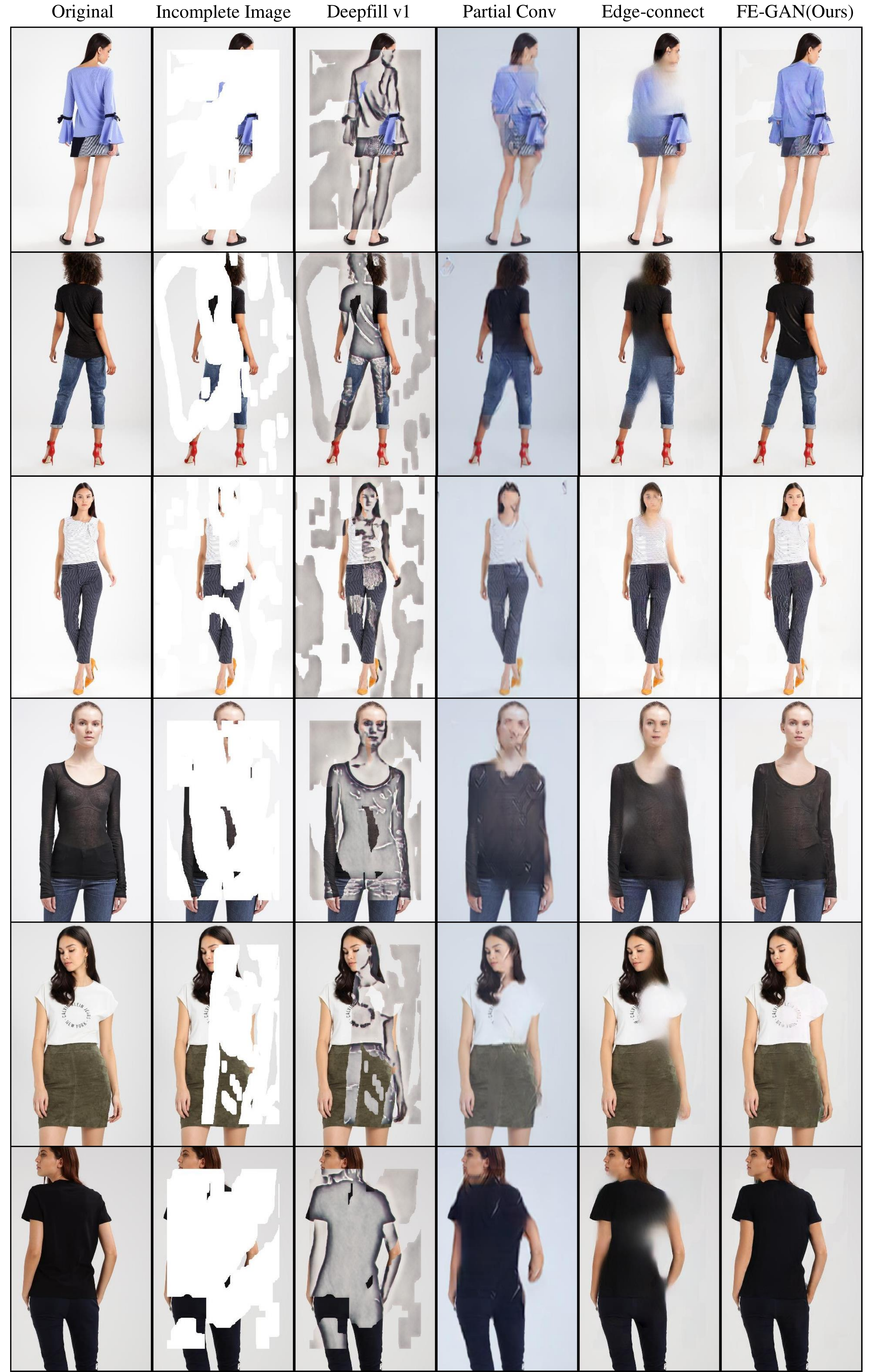} 
    \caption{Qualitative comparisons between Deepfill v1~\cite{Yu2018GenerativeII}, Partial Conv~\cite{Liu2018ImageIF}, Edge-connect~\cite{Nazeri2019EdgeConnectGI}, and FE-GAN on MPV~\cite{dong2019towards}.}
    \label{fig:mpv}
\end{figure}

\end{document}